\documentclass[11pt,a4paper]{article}
\usepackage{authblk}
\usepackage{stackengine}
\usepackage[hyperref]{emnlp2018}
\aclfinalcopy
\usepackage{times}
\usepackage{latexsym}

\usepackage{url}
\usepackage{xcolor}
\usepackage{upgreek}
\usepackage[shortlabels]{enumitem}
\usepackage{tabu}
\setlist[enumerate]{nosep}
\usepackage{floatrow}
\usepackage{multirow}
\usepackage{makecell}
\usepackage{amsmath}

\usepackage{booktabs}
\usepackage{graphicx}
\usepackage{amsfonts}
\usepackage{tabularx}
\usepackage{bm}

\def\aclpaperid{***}
\DeclareFloatFont{tiny}{\tiny}
\usepackage{silence}
\usepackage{tabularx}
\usepackage{pifont}
\usepackage{booktabs}
\usepackage{multirow}
\usepackage{subfig}
\usepackage{hyperref}
\usepackage{pifont}
\usepackage{xcolor}
\usepackage{amssymb}
 
\usepackage[scr=dutchcal]{mathalpha}

\newlength\Origarrayrulewidth

\usepackage{amsmath} 
\usepackage{wrapfig}

\renewcommand{\footnotesize}{\fontsize{8pt}{6pt}\selectfont}

\usepackage{graphicx}
\graphicspath{ {./images/} }
\usepackage[ruled]{algorithm2e}
\usepackage{amsthm}
\usepackage{amssymb}

\makeatletter
\newcommand{\vo}{\vec{o}\@ifnextchar{^}{\,}{}}
\makeatother

\title{VOGUE: Answer Verbalization through Multi-Task Learning}

\author{Endri Kacupaj$^{1}$, Shyamnath Premnadh$^{1}$, Kuldeep Singh$^{2,3}$, Jens Lehmann$^{1,4}$ \\
\textbf{Maria Maleshkova$^1$} \\
$^1$University of Bonn,
$^2$Zerotha Research,
$^3$Cerence GmbH,
$^4$Fraunhofer IAIS\\
{\tt \{kacupaj,jens.lehmann,maleshkova\}@cs.uni-bonn.de}\\
{\tt s6shprem@uni-bonn.de}\\
{\tt kuldeep.singh1@cerence.com}\\
{\tt jens.lehmann@iais.fraunhofer.de}
}

\date{}

\begin{document}
\setlength{\abovedisplayskip}{3pt}
\setlength{\belowdisplayskip}{3pt}

\maketitle

\begin{abstract}
In recent years, there have been significant developments in Question Answering over Knowledge Graphs (KGQA). Despite all the notable advancements, current KGQA systems only focus on answer generation techniques and not on answer verbalization. However, in real-world scenarios (e.g., voice assistants such as Alexa, Siri, etc.), users prefer verbalized answers instead of a generated response. This paper addresses the task of answer verbalization for (complex) question answering over knowledge graphs. In this context, we propose a multi-task-based answer verbalization framework: VOGUE (\textbf{V}erbalization thr\textbf{O}u\textbf{G}h m\textbf{U}lti-task l\textbf{E}arning). The VOGUE framework attempts to generate a verbalized answer using a hybrid approach through a multi-task learning paradigm. Our framework can generate results based on using questions and queries as inputs concurrently. VOGUE comprises four modules that are trained simultaneously through multi-task learning. We evaluate our framework on all existing datasets for answer verbalization, and it outperforms all current baselines on both BLEU and METEOR scores as evaluation metric.
\end{abstract}


\section{Introduction}
In recent years, publicly available knowledge graphs (KG) (e.g., DBpedia~\cite{dbpedia}, Wikidata~\cite{wikidata}) have been broadly adopted as a source of knowledge in several tasks such as entity linking, relation extraction, and question answering~\cite{kacupaj2021lasagne}. Question answering (QA) over knowledge graphs, in particular, is an essential task that maps a user's utterance to a query over a KG to retrieve the correct answer~\cite{singh2018reinvent}. The initial knowledge graph question answering systems (KGQA) were mostly template- or rule-based systems with limited learnable modules~\cite{unger2012template}. With the increasing popularity of intelligent personal assistants (e.g., Alexa, Siri), the research focus has been shifted to conversational question answering over KGs (ConvQA) that involve single-turn/multi-turn dialogues~\cite{kacupaj2021paraqa}. 

\begin{figure}[t]
\centering
\includegraphics[height=0.45\textwidth]{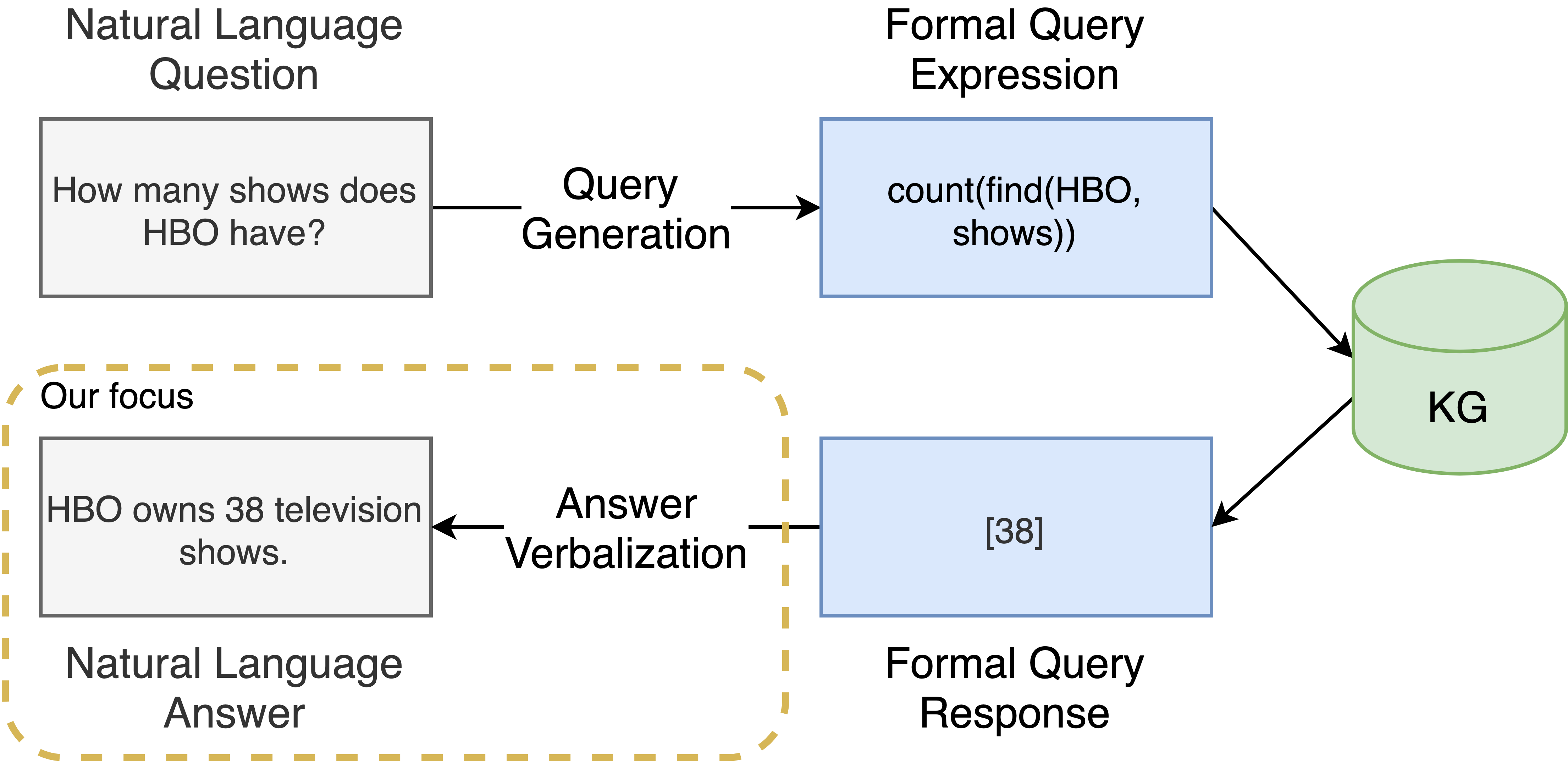}
\caption{A QA pipeline with integrated answer verbalization module. Our focus is the answer verbalization task as we assume logical form is generated by a QA system using the input question.}
\label{fig:into}
\end{figure}

Existing open-source KGQA systems are restricted to only generating or producing answers without verbalizing them in natural language~\cite{fu2020survey}. The lack of verbalization makes the interaction with the user not natural in contrast to voice assistants such as Siri and Alexa. Figure~\ref{fig:into} depicts an ideal integration of a QA pipeline with answer verbalization. For instance, assuming that the answer to the exemplary question, ``How many shows does HBO have?" is not known by the user. Suppose the QA system only responds with a number (e.g., 38) as an answer (similar to open-source KGQA systems), with no further explanation. In that case, the user might need to refer to an external data source to verify the answer. 
In an attempt to enable the users to verify the answer provided by a QA system, researchers employed techniques such as (i) revealing the generated formal query~\cite{ferre2017sparklis}, (ii) graphical visualizations of the formal query~\cite{zheng2017natural} and (iii) verbalizing the formal query~\cite{ell2014sparql}. Understanding the necessity of verbalized answers in the KGQA domain, recently, several datasets have been proposed ~\cite{kacupaj2020vquanda,biswas2021vanilla}. For the answer verbalization task, the system has to verbalize the answer to convey not only the information requested by the user but also additional characteristics that indicate how the answer was determined. In our exemplary question (from \citet{kacupaj2021paraqa} dataset), a verbalized response would look like, ``HBO owns 38 television shows." or ``There are 38 TV shows whose owner is HBO.". Both answers allow the user to verify that the system retrieved the total number of TV shows owned by HBO. In the literature, there exist empirical results showing that answer verbalization quantitatively and qualitatively improves the ability to understand the answer~\cite{kacupaj2021paraqa}. However, it remains an open question -- How can we verbalize an answer, given a logical form and an input question. With our work we address precisely this open and highly relevant research question.

In this paper, we propose VOGUE, the first approach dedicated to verbalize answers for KGQA. Our idea is to employ the question (user utterance) and the QA system-generated query as inputs. We refer to this strategy as ``hybrid", since the final verbalized answer is produced using both the question and query concurrently. This work argues that leveraging content from both sources allows the model for better convergence and provides new, improved results. Furthermore, we complement our hypothesis with multi-task learning paradigms, since multi-task learning has been quite efficient for different system architectures~\cite{cipolla2018mltloss}, including question answering systems~\cite{plepi2021carton,kacupaj2021lasagne}.
Our framework can receive two (e.g., question \& query) or even one (e.g., question) input. It consists of four modules that are trained simultaneously to generate the verbalized answer. The first module employs a dual transformer-based encoder architecture for encoding the inputs. The second module determines whether the encoded inputs are relevant and decides if both will be used for verbalization. The third module consists of a cross-attention network that performs question and query matching by jointly modeling the relationships of question words and query actions. The final module employs a transformer decoder that is used to generate the final verbalization.
Our work makes the following contributions:
\begin{itemize}
    \item  We introduce the first multi-task-based hybrid answer verbalization framework consisting of simultaneously trained four modules.
    \item We propose a similarity threshold and cross attention modules to determine the relevance between the inputs and fuse information to employ a hybrid strategy.
    \item We provide an extensive evaluation and ablation study of the proposed framework on three QA datasets with answer verbalization. Our evaluation results establish a new baseline for answer verbalization task, which we believe will drive future research in a newly studied problem.
\end{itemize}
\noindent To facilitate reproducibility and reuse, our framework implementation is publicly available\footnote{\url{https://github.com/endrikacupaj/VOGUE}}. The structure of the paper is as follows: Section 2 summarizes the related work. Section 3 provides the task definition. Section 4 presents the proposed framework. Section 5 describes the experiments, results, ablation study and error analysis. We conclude in Section 6.

\section{Related Work}
As part of the related work we describe previous efforts and refer to different approaches from research fields, including task-oriented dialog systems, WebNLG, and KGQA systems.

A task-oriented dialogue system aims to help the user complete certain tasks in a specific domain (e.g. restaurant booking, weather query, or flight booking), making it valuable for real-world business. Typically, task-oriented dialogue systems are built on top of a structured ontology, which defines the tasks' domain knowledge.
Work in~\cite{bordes2017god} formalized the task-oriented dialogue as a reading comprehension task regarding the dialogue history as context, user utterance as the question, and system response as the answer. In their work, authors utilized end-to-end memory networks for multi-turn inference. 
In~\cite{lei-etal-2018-sequicity}, authors proposed a two-step seq2seq generation model, which bypassed the structured dialogue act representation and only retain the dialogue state representation. 
~\citet{kassawat2019multi} proposed RNN-based end-to-end encoder-decoder architecture, which employs joint embeddings of the knowledge graph and the corpus as input. 

The WebNLG is a challenge that consists of mapping structured data to a textual representation. 
The dataset~\cite{gardent-etal-2017-creating} contains data/text pairs where the data is a set of triples extracted from DBpedia, and the text is the verbalization of these triples. The dataset has been promoted for the development of 1) RDF verbalizers and 2) microplanners to handle a wide range of linguistic constructions. In our case, we only focus on related work pertaining to RDF verbalizers. 
~\citet{zhao-etal-2020-bridging} propose DualEnc, a dual encoding model that can incorporate the graph structure and cater to the linear structure of the output text. \citet{song-etal-2020-structural} proposes a graph-to-text approach that leverages richer training signals to guide the model for preserving input information. They introduce two types of autoencoding losses, each individually focusing on different aspects of input graphs. The losses are then back-propagated to calibrate the model via multi-task training. Work in \citet{ijcai2020-524} propose an attention-based model, which mainly contains an entity extraction module and a relation detection module. The model devises a supervised multi-head self-attention mechanism as the relation detection module to learn the token-level correlation for each relation type separately.


The task-oriented dialogue systems and WebNLG contain various approaches for generating text; nevertheless, none of them can be applied directly to solve answer verbalization for KGQA systems. Most task-oriented dialogue systems are designed and implemented to fit their corresponding task, and therefore they would not be suitable for open-domain knowledge graphs (e.g. Wikidata, DBpedia). Regarding WebNLG, the task only considers triples or graph structure data as input. In answer verbalization, the model input can be the question and/or the query. While the query can be translated into a graph structure, there is no support for textual information such as the question. 

\section{Task Definition}
In this work, we target the problem of answer verbalization for KGQA. A semantic parsing-based QA system maps the question into executable logical forms and then executes it on a KG to produce the answer.
For our task, given the question, the generated logical form, and the extracted answer, we aim to generate a natural language sentence, with the requirements that it is grammatically sound and correctly represents all the information in the question, logical form, and answer. 
Formally, let $X, Y$ denote the source-target pair. $X$ contains the set of questions, logical forms, answers, and $Y$ corresponds to $y_1,y_2,...,y_m$, which is the verbalized answer of $X$. The goal of the answer verbalization is to learn a distribution $p(Y|X)$ to generate natural language text describing the answer automatically.

\section{Approach}
\begin{figure*}[!ht]
\includegraphics[height=0.7\textwidth]{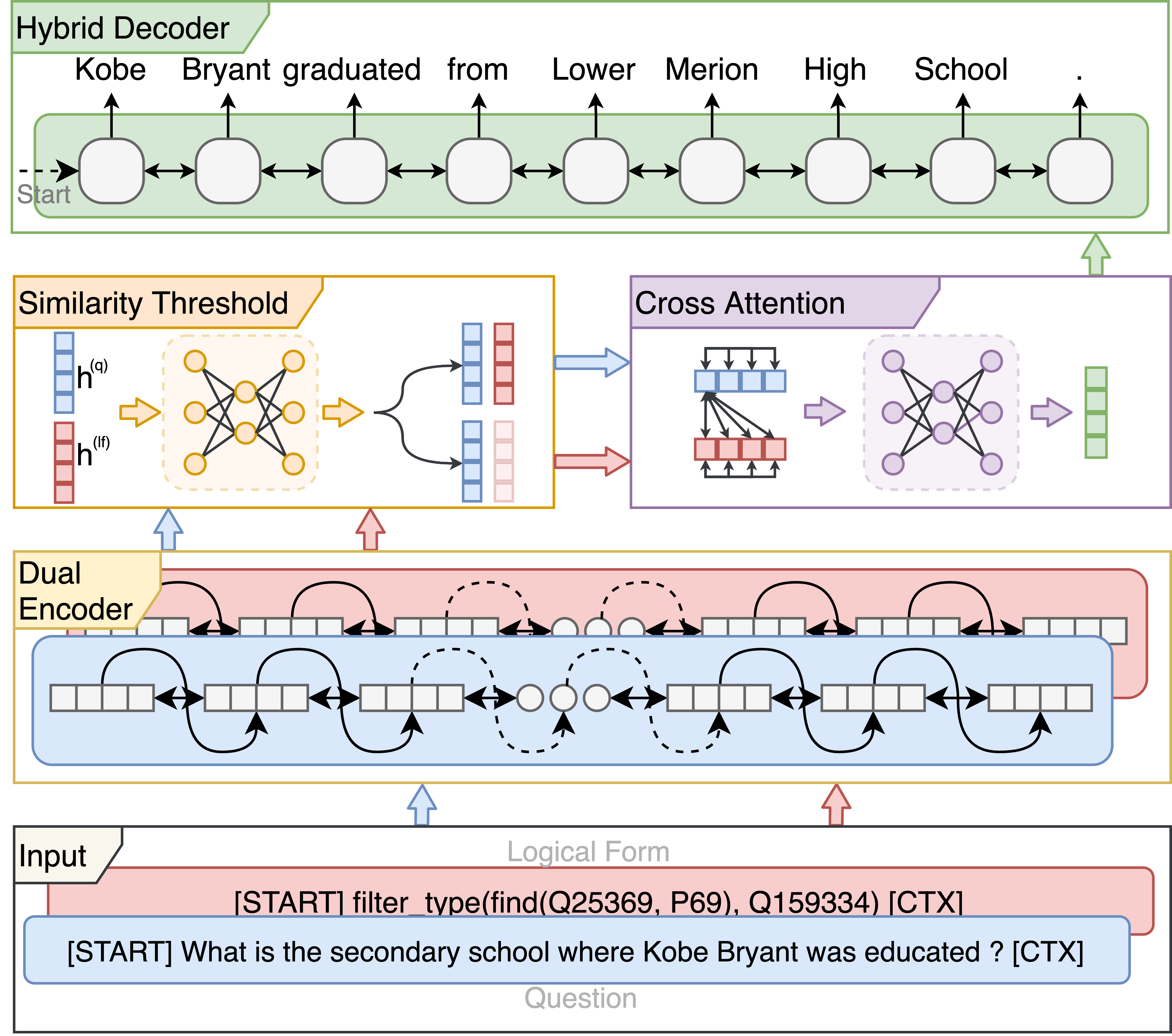}
\caption{VOGUE's architecture. It consists of four modules: 1) A dual encoder that is responsible to encode both inputs (question, logical form). 2) A similarity threshold module that determines whether the encoded inputs are relevant and determines if both will be used for verbalization. 3) A cross-attention module that performs question and query matching by jointly modeling the relationships of question words and query actions. 4) A hybrid decoder that generates the verbalized answer using the information of both question and logical form representations from the cross-attention module.}
\label{fig:model_architecture}
\end{figure*}

In question answering, the input data consists of question $u$ and its answer $a$, extracted from the knowledge graph. The QA system will map the question to a logical form $z$ depending on the context. For answer verbalization, VOGUE maps the question, logical form, and answer to natural language sentence $s$. Figure~\ref{fig:model_architecture} shows the architecture of VOGUE.

\subsection{Dual Encoder}
To encode both the question and logical form, we employ a dual encoder architecture. Our dual encoder consists of two instances of the transformer encoder~\cite{vaswani2017attention}. 

First, as a preprocessing, we use a previous competitive pre-trained named entity recognition model~\cite{yamada-etal-2020-luke} to identify and replace all entities in the question with a more general entity token $[ENT]$. In this way, we allow our model to focus on the sentence structure and relations between words. Furthermore, our model learns the positions of entities in the question. It also allows VOGUE to predict the respective entity positions in the verbalized answer. The same preprocessing step applies to the logical form. At the end of each input, we append a context token $[CTX]$, which is used later as a semantic representation.

Next, given the question utterance $q$ containing $n$ words $\{w_1,\dots,w_n\}$ and the logical form $l$ containing $m$ actions $\{a_1,\dots,a_m\}$, we tokenize the contexts and use the pre-trained model GloVe~\cite{pennington-etal-2014-glove} to embed the words into a vector representation space of dimension $d$ \footnote{We employ the same dimension $d$ for all the representations, unless it is explicitly mentioned.}. Our word embedding model provides us with the sequences $x^{(q)} = \{x^{(q)}_1,\dots,x^{(q)}_n\}$, $x^{(lf)} = \{x^{(lf)}_1,\dots,x^{(lf)}_m\}$ where $x^{(q)}_i$, $x^{(lf)}_i$ are given by,
\begin{equation}
    \begin{split}
        &x^{(q)}_i = GloVe(w_i), \\
        &x^{(lf)}_i = GloVe(a_i),
    \end{split}
\end{equation}
\noindent and $x^{(q)}_i, x^{(lf)}_i \in \mathbb{R}^{d}$. Afterwards, both sequences are forwarded through the transformer encoders. The two encoders here output the contextual embeddings $h^{(q)} = \{h_{1}^{(q)},\dots,h_{n}^{(q)}\}$ and $h^{(lf)} = \{h_{1}^{(lf)},\dots,h_{m}^{(lf)}\}$, where $h_{i}^{(q)}, h_{i}^{(lf)} \in \mathbb{R}^{d}$. We define this as: 
\begin{equation}
\begin{split}
    &h^{(q)} = encoder_{q}(x_{q}; \theta^{(enc_{q})}), \\
    &h^{(lf)} = encoder_{lf}(x_{lf}; \theta^{(enc_{lf})}),
\end{split}
\end{equation}
\noindent where $\theta^{(enc_{q})}$, $\theta^{(enc_{lf})}$ are the encoders trainable parameters.

\subsection{Similarity Threshold}
Given the encoded question utterance and logical form, VOGUE's second module is responsible for learning the relevance between the inputs and determining whether we will employ both for verbalization. This module is necessary when we want to utilize our framework alongside a question answering system. If we assume that the QA system is perfect and always produces correct logical forms, this module can be skipped. However, in a real-world scenario, QA systems are far from perfect. Therefore, we employ this module, which intends to identify the threshold for determining if two inputs are similar or not. The input here is the concatenation of the hidden states of the encoded question utterance $h^{(q)}$ and logical form $h^{(lf)}$. The module will perform binary classification on the vocabulary $V^{(st)} = \{0, 1\}$, where $0$ indicates that there is no high relevance between the inputs, and only the question will be used for verbalization. While $1$ allows us to use both and continue with the next module.
Overall, our similarity threshold module is implemented using two linear layers, a Leaky ReLU activation function and a softmax for the predictions. Formally we define the module as:

\begin{equation}
\begin{split}
    &h^{(st)} = LeakyReLU( \boldsymbol{W}^{(st_1)} [h^{(q)};h^{(lf)}]), \\
    &p^{(st)} = softmax( \boldsymbol{W}^{(st_2)} h^{(st)}),
\end{split}
\end{equation}

\noindent where $\boldsymbol{W}^{(st_1)} \in \mathbb{R}^{d \times 2d}$ are the weights of the first linear layer and $h^{(st)}$ is the hidden state of the module. $\boldsymbol{W}^{(st_2)} \in \mathbb{R}^{|V^{(st)}| \times d}$ are the weights of the second linear layer, $|V^{(st)}|$ is the size of the vocabulary and $p^{(st)}$ denotes the probability distribution over the vocabulary indices.

\subsection{Cross Attention}
Inspired by recent computer vision research~\cite{wei2020crossattention,mohla2020crossattention}, we employ a cross-attention module that exploits relationships between the inputs and fuses information. The module here performs question and logical form matching by jointly modeling the relationships of question words and logical form actions. Our cross-attention approach is a variation of the self-attention mechanism~\cite{vaswani2017attention}. In the self-attention mechanism the output is determined by a query and a set of key-value pairs. Given the stacked encoded question and logical form, $h^{(qlf)} = \binom{h^{(q)}}{h^{(lf)}} = \{h_{1}^{(q)},\dots,h_{n}^{(q)};h_{1}^{(lf)},\dots,h_{m}^{(lf)}\}$, where $h^{(qlf)} \in \mathbb{R}^{2 \times d}$ we calculate the query and key-value pairs using three linear projections:

\begin{equation}
    \small
    \begin{split}
        &\boldsymbol{Q}^{(qlf)} = \boldsymbol{W}^{(Q)} h^{(qlf)} = \binom{\boldsymbol{W}^{(Q)}h^{(q)}}{\boldsymbol{W}^{(Q)}h^{(lf)}} = \binom{\boldsymbol{Q}^{(q)}}{\boldsymbol{Q}^{(lf)}}, \\
        &\boldsymbol{K}^{(qlf)} = \boldsymbol{W}^{(K)} h^{(qlf)} = \binom{\boldsymbol{W}^{(K)}h^{(q)}}{\boldsymbol{W}^{(K)}h^{(lf)}} = \binom{\boldsymbol{K}^{(q)}}{\boldsymbol{K}^{(lf)}}, \\
        &\boldsymbol{V}^{(qlf)} = \boldsymbol{W}^{(V)} h^{(qlf)} = \binom{\boldsymbol{W}^{(V)}h^{(q)}}{\boldsymbol{W}^{(V)}h^{(lf)}} = \binom{\boldsymbol{V}^{(q)}}{\boldsymbol{V}^{(lf)}},
    \end{split}
\end{equation}

\noindent where $\boldsymbol{W}^{(Q)}, \boldsymbol{W}^{(K)}, \boldsymbol{W}^{(V)} \in \mathbb{R}^{d \times d}$ are the weights of the linear layers and $\boldsymbol{Q}^{(qlf)}, \boldsymbol{K}^{(qlf)}, \boldsymbol{V}^{(qlf)}$ are the query, key and value of the stacked question and logical form. Next, for calculating the cross-attention we simplify the ``\textit{Scaled Dot-Product Attention}"~\cite{vaswani2017attention} step by removing the scaling factor and softmax. We end-up calculating the attention of our input as described below:

\begin{equation}
    \small
    \begin{split}
        Attention&(\boldsymbol{Q}^{(qlf)}, \boldsymbol{K}^{(qlf)}, \boldsymbol{V}^{(qlf)})\\ \; &= \; \boldsymbol{Q}^{(qlf)}\boldsymbol{K}^{(qlf)T} \cdot \boldsymbol{V}^{(qlf)} \\
        &= \; \binom{\boldsymbol{Q}^{(q)}}{\boldsymbol{Q}^{(lf)}} (\boldsymbol{K}^{(q)T} \boldsymbol{K}^{(lf)T})\cdot \binom{\boldsymbol{V}^{(q)}}{\boldsymbol{V}^{(lf)}} \\
        &= \; \binom{\boldsymbol{Q}^{(q)}\boldsymbol{K}^{(q)T} \;\;\;\; \boldsymbol{Q}^{(q)}\boldsymbol{K}^{(lf)T}}{\boldsymbol{Q}^{(lf)}\boldsymbol{K}^{(q)T} \;\;\;\; \boldsymbol{Q}^{(lf)}\boldsymbol{K}^{(lf)T}} \cdot \binom{\boldsymbol{V}^{(q)}}{\boldsymbol{V}^{(lf)}} \\
        &= \; \binom{\boldsymbol{Q}^{(q)}\boldsymbol{K}^{(q)T}\boldsymbol{V}^{(q)} \; + \; \boldsymbol{Q}^{(q)}\boldsymbol{K}^{(lf)T}\boldsymbol{V}^{(lf)}}{\boldsymbol{Q}^{(lf)}\boldsymbol{K}^{(lf)T}\boldsymbol{V}^{(lf)} \; + \; \boldsymbol{Q}^{(lf)}\boldsymbol{K}^{(q)T}\boldsymbol{V}^{(q)}}.
    \end{split}
\end{equation}

\noindent While calculating the cross-attention for the question, we also use the key-value pair from the logical form ($\boldsymbol{K}^{(lf)}, \boldsymbol{V}^{(lf)}$), the same applies when calculating the cross-attention for the logical form. After calculating the cross-attentions, we use the same steps as in the transformer to produce the new representations for our inputs. Finally, considering $h^{(qca)}, h^{(lfca)}$ the output representations of the cross-attention module for the question and logical form respectively, we concatenate them and forward them to the hybrid decoder module.

\subsection{Hybrid Decoder}
To translate the input question and logical form into a sequence of words (verbalized answer), we utilize a transformer decoder architecture~\cite{vaswani2017attention}, which employs the multi-head attention mechanism. The decoder will generate the final natural language answer. The output here is dependent on the cross-attention embedding $h^{(ca)}$. Here we define the decoder vocabulary as 

\begin{equation}
    \begin{split}
    V^{(dec)} = V^{(vt)} \; \cup \; \{\; &[START], \; [END],\\ \; &[ENT], \; [ANS] \; \},
    \end{split}
\end{equation}

\noindent where $V^{(vt)}$ is the vocabulary with all the distinct tokens from our verbalizations. As we can see, the decoder vocabulary contains four additional helper tokes, where two of them ($[START]$, $[END]$) indicate when the decoding process starts and ends, while the other two ($[ENT]$, $[ANS]$) are used to specify the position of the entities and the answer on the final verbalized sequence. On top of the decoder stack, we employ a linear layer alongside a softmax to calculate each token's probability scores in the vocabulary. We define the decoder stack output as follows:

\begin{equation}
\begin{split}
    &h^{(dec)} = decoder(h^{(ca)};\theta^{(dec)}),\\
    &p_{t}^{(dec)} = softmax(\boldsymbol{W}^{(dec)} h_{t}^{(dec)}),
\end{split}
\end{equation}

\noindent where $h_{t}^{(dec)}$ is the hidden state in time step $t$, $\theta^{(dec)}$ are the decoder trainable parameters, $\boldsymbol{W}^{(dec)} \in \mathbb{R}^{|V^{(dec)}|\times 2d}$ are the linear layer weights, and $p_{t}^{(dec)} \in \mathbb{R}^{|V^{(dec)}|}$ is the probability distribution over the decoder vocabulary in time step $t$. The $|V^{(dec)}|$ denotes the vocabulary size of the decoder module in VOGUE.

\subsection{Learning}
The framework consists of four trainable modules. However, we apply a loss function only on two of them (similarity threshold and hybrid decoder). The dual encoder and cross-attention modules are trained based on the similarity threshold and hybrid decoder's signal.
To account for multi-tasking, we perform a weighted average of all the single losses:

\begin{equation}
    L = \lambda_1 L^{st} + \lambda_2 L^{dec},
\end{equation}

\noindent where $\lambda_1, \lambda_2$ are the relative weights learned during training considering the difference in magnitude between losses by consolidating the log standard deviation \cite{armitage2020mlm,cipolla2018mltloss}. $L^{st}$ and $L^{dec}$ are the respective negative log-likelihood losses of the similarity threshold and hybrid decoder modules. These losses are defined as: 

\begin{equation}
\begin{split}
    &L^{st} = - \sum_{j=1}^{2d} log p(y_{j}^{(st)} | x), \\
    &L^{dec} = - \sum_{k=1}^{m} log p(y_{k}^{(dec)} | x),
\end{split}
\end{equation}

\noindent where $m$ is the length of the gold logical form. 
$y_{j}^{(st)} \in V^{(st)}$ are the gold labels for the similarity threshold and
$y_{k}^{(dec)} \in V^{(dec)}$ are the gold labels for the decoder. The model benefits from each module's supervision signals, which improves the performance in the given task.

\section{Experiments}
\subsection{Experimental Setup}

\begin{table}
\centering
\caption{Dataset statistics, including the (average) number of tokens per question sentence, the (average) number of tokens per answer sentence and the vocabulary list size.}
\label{tab:dataset_stats}
\resizebox{\columnwidth}{!}{%
\begin{tabular}{l|ccccc}
\toprule
\textbf{Dataset} & \textbf{Train} & \textbf{Test} & \textbf{Ques len.} & \textbf{Ans len.} & \textbf{Vocab.} \\ \midrule
VQuAnDa & 4000 & 1000 & 12.27 & 16.95 & 10431 \\
ParaQA & 12637 & 3177 & 12.27 & 17.06 & 12755 \\
VANiLLa & 85729 & 21433 & 8.96 & 8.98 & 50505 \\
\bottomrule
\end{tabular}%
}
\end{table}

\begin{table*}
\centering
\def\arraystretch{1.2}
\caption{Results on answer verbalization. VOGUE outperforms all existing baselines and achieves the new state of the art for both the BLEU and METEOR scores. The baseline experiment results are reported with two inputs: Question (\textbf{Q}) and gold Logical Form (\textbf{LF}), while VOGUE employs a Hybrid (\textbf{H}) approach.}
\label{tab:results}
\begin{tabular}{lcccccc}
\toprule
 & \multicolumn{3}{c|}{\textbf{BLEU}} & \multicolumn{3}{c}{\textbf{METEOR}} \\ \cmidrule{2-7} 
\multicolumn{1}{c}{\textbf{Models}} & \multicolumn{1}{c|}{\textbf{VQuAnDa}} & \multicolumn{1}{c|}{\textbf{ParaQA}} & \multicolumn{1}{c|}{\textbf{VANiLLa}} & \multicolumn{1}{c|}{\textbf{VQuAnDa}} & \multicolumn{1}{c|}{\textbf{ParaQA}} & \textbf{VANiLLa} \\ \midrule
\multicolumn{1}{l|}{RNN~\cite{luong2015effective} (\textbf{Q})} & 15.43 & 22.45 & \multicolumn{1}{c|}{16.66} & 53.15 & 58.41 & 58.67 \\
\multicolumn{1}{l|}{RNN~\cite{luong2015effective} (\textbf{LF})} & 20.19 & 26.36 & \multicolumn{1}{c|}{16.45} & 57.06 & 61.87 & 55.34 \\
\multicolumn{1}{l|}{Convolutional~\cite{gehring2017conv} (\textbf{Q})} & 21.32 & 25.94 & \multicolumn{1}{c|}{15.42} & 57.54 & 60.82 & 61.14 \\
\multicolumn{1}{l|}{Convolutional~\cite{gehring2017conv} (\textbf{LF})} & 26.02 & 31.89 & \multicolumn{1}{c|}{16.89} & 64.30 & 65.85 & 58.72 \\
\multicolumn{1}{l|}{Transformer~\cite{vaswani2017attention} (\textbf{Q})} & 18.37 & 23.61 & \multicolumn{1}{c|}{30.80} & 56.83 & 59.63 & 62.16 \\
\multicolumn{1}{l|}{Transformer~\cite{vaswani2017attention} (\textbf{LF})} & 23.18 & 28.01 & \multicolumn{1}{c|}{28.12} & 60.17 & 63.75 & 59.01 \\
\multicolumn{1}{l|}{BERT~\cite{devlin-etal-2019-bert} (\textbf{Q})} & 22.78 & 26.12 & \multicolumn{1}{c|}{31.32} & 59.28 & 62.59 & 62.96 \\
\multicolumn{1}{l|}{BERT~\cite{devlin-etal-2019-bert} (\textbf{LF})} & 26.48 & 30.31 & \multicolumn{1}{c|}{30.11} & 65.92 & 65.92 & 59.27 \\ \midrule
\multicolumn{1}{l|}{VOGUE (Ours) (\textbf{H})} & \textbf{28.76} & \textbf{32.05} & \multicolumn{1}{c|}{\textbf{35.46}} & \textbf{67.21} & \textbf{68.85} & \textbf{65.04} \\
\bottomrule
\end{tabular}
\end{table*}

\paragraph{\textbf{Datasets}}
We perform experiments on three answer verbalization datasets (cf., Table~\ref{tab:dataset_stats}). Below we provide a brief description of these:
\begin{itemize}
    \item VQuAnDa~\cite{kacupaj2020vquanda} is the first QA dataset, which provides the verbalization of the answer in natural language. It contains $5000$ ``\textit{complex}" questions with their SPARQL queries and answers verbalization. The dataset consists of $5042$ entities and $615$ relations.
    \item ParaQA~\cite{kacupaj2021paraqa} is a QA dataset with multiple paraphrased responses. The dataset was created using a semi-automated framework for generating diverse paraphrasing of the answers using techniques such as back-translation. It contains $5000$ ``\textit{complex}" question-answer pairs with a minimum of two and a maximum of eight unique paraphrased responses for each question.
    \item VANiLLa~\cite{biswas2021vanilla} is a QA dataset that offers answers in natural language sentences. The answer sentences in this dataset are syntactically and semantically closer to the question than the triple fact. The dataset consists of over $100k$ ``\textit{simple}" questions.
\end{itemize}
\textbf{\textbf{Model Configuration}}
For simplicity, to represent the logical forms, we employ the same grammar as in~\cite{kacupaj2021lasagne}. Our approach can be used with any other grammar or even directly with SPARQL queries. However, we believe it is better to employ semantic grammar from a state-of-the-art QA model. To properly train the similarity threshold module, we had to introduce negative logical forms for each question. We did that by corrupting the gold logical forms, either by replacing a random action or finding another ``\textit{similar}" logical form from the dataset based on the Levenshtein distance. For all the modules in our framework, we employ an embedding dimension of $300$. A transformer encoder and decoder having two layers and six heads for the multi-head attention model is used. We apply dropout~\cite{srivastava2014dropout} with a probability $0.1$. For the optimization, we use the Noam optimizer proposed by~\cite{vaswani2017attention}, where authors use an Adam optimizer~\cite{kingma2015adam} with several warmup steps for the learning rate.\\
\textbf{Model for Comparison}
We compare our framework with the four baselines that have been evaluated on the considered datasets. All baselines consist of sequence to sequence architectures, a family of machine learning approaches used for language processing and often used for natural language generation tasks. The first model consists of an RNN~\cite{luong2015effective} based architecture, the second uses a convolutional network~\cite{gehring2017conv}, the third employs a transformer network~\cite{vaswani2017attention}, while the last one uses pre-trained BERT~\cite{devlin-etal-2019-bert} model. 
For a fair comparison with our framework, we report the baselines' results using the question and the logical form as separate inputs considering that baselines are limited to accept both inputs together.\\
\textbf{Evaluation Metrics}
We use the same metrics as employed by the authors of the three existing datasets \cite{kacupaj2020vquanda,kacupaj2021paraqa,biswas2021vanilla} on the previously mentioned baselines. The BLEU score, as defined by~\cite{papineni2002bleu}, analyzes the co-occurrences of n-grams in the reference and the proposed responses. It computes the n-gram precision for the whole dataset, which is then multiplied by a brevity penalty to penalize short translations. We report results for BLEU-4. The METEOR score introduced by~\cite{banerjee2005meteor} is based on the harmonic mean of uni-gram precision and recall, with recall weighted higher than precision. 

\begin{table*}[t!]
\centering
\def\arraystretch{1.2}
\caption{Results of the answer verbalization with a semantic parsing QA system. VOGUE still outperforms all baselines. For the baselines we employ only the question as input, while our framework employs the similarity threshold module to determine whether a hybrid verbalization can be performed.}
\label{tab:results_with_qa}
\begin{tabular}{lcccccc}
\toprule
 & \multicolumn{3}{c|}{\textbf{BLEU}} & \multicolumn{3}{c}{\textbf{METEOR}} \\ \cmidrule{2-7} 
\multicolumn{1}{c}{\textbf{Models}} & \multicolumn{1}{c|}{\textbf{VQuAnDa}} & \multicolumn{1}{c|}{\textbf{ParaQA}} & \multicolumn{1}{c|}{\textbf{VANiLLa}} & \multicolumn{1}{c|}{\textbf{VQuAnDa}} & \multicolumn{1}{c|}{\textbf{ParaQA}} & \textbf{VANiLLa} \\ \midrule
\multicolumn{1}{l|}{RNN} & 15.43 & 22.45 & \multicolumn{1}{c|}{16.66} & 53.15 & 58.41 & 58.67 \\
\multicolumn{1}{l|}{Convolutional} & 21.32 & 25.94 & \multicolumn{1}{c|}{15.42} & 57.54 & 60.82 & 61.14 \\
\multicolumn{1}{l|}{Transformer} & 18.37 & 23.61 & \multicolumn{1}{c|}{30.80} & 56.83 & 59.63 & 62.16 \\
\multicolumn{1}{l|}{BERT} & 22.78 & 26.12 & \multicolumn{1}{c|}{31.32} & 59.28 & 62.59 & 62.96 \\ \midrule
\multicolumn{1}{l|}{VOGUE (Ours)} & \textbf{25.76} & \textbf{28.42} & \multicolumn{1}{c|}{\textbf{33.14}} & \textbf{64.61} & \textbf{67.52} & \textbf{63.69} \\
\bottomrule
\end{tabular}
\end{table*}

\begin{table*}[t!]
\centering
\def\arraystretch{1.2}
\caption{Ablation study results that indicate the effectiveness of cross attention and multi-task learning. The first row contains the results of the VOGUE framework when training all four modules with multi-task learning. The second and third rows selectively remove the cross attention and the multi-task learning from VOGUE. Best values in bold.} 
\label{tab:ablation}
\begin{tabular}{lcccccc}
\toprule
 & \multicolumn{3}{c|}{\textbf{BLEU}} & \multicolumn{3}{c}{\textbf{METEOR}} \\ \cmidrule{2-7} 
\multicolumn{1}{c}{\textbf{Ablation}} & \multicolumn{1}{c|}{\textbf{VQuAnDa}} & \multicolumn{1}{c|}{\textbf{ParaQA}} & \multicolumn{1}{c|}{\textbf{VANiLLa}} & \multicolumn{1}{c|}{\textbf{VQuAnDa}} & \multicolumn{1}{c|}{\textbf{ParaQA}} & \textbf{VANiLLa} \\ \midrule
\multicolumn{1}{l|}{Ours} & \textbf{28.76} & \textbf{32.05} & \multicolumn{1}{c|}{\textbf{35.46}} & \textbf{67.21} & \textbf{68.85} & \textbf{65.04} \\
\multicolumn{1}{l|}{w/o Cross Attention} & 26.24 & 30.59 & \multicolumn{1}{c|}{30.94} & 64.93 & 66.16 & 62.12 \\
\multicolumn{1}{l|}{w/o Multi-Task Learning} & 25.74 & 28.15 & \multicolumn{1}{c|}{29.07} & 62.31 & 63.84 & 61.49 \\
\bottomrule
\end{tabular}
\end{table*}

\subsection{Results}

Table~\ref{tab:results} summarizes the results comparing the VOGUE framework to the previous baselines for answer verbalization. VOGUE significantly outperforms the earlier baselines for both the BLEU and METEOR scores. While for the other baselines, we perform experiments with two different inputs (Question, gold Logical Form), VOGUE is the only one that directly uses both inputs (Hybrid). As we can see, for both datasets VQuAnDa and ParaQA, all baselines perform slightly worse when receiving the question as input compared to the gold logical form. This is due to the constant input pattern templates that the logical forms have. However, this does not apply to the VANiLLa dataset since it only contains simple questions. VOGUE achieves a BLEU score of $28.76$ on VQuAnDa, which is $2$ points higher than the second-best BERT (LF). The same applies to the METEOR score.
Regarding ParaQA, VOGUE performs slightly better than the second Convolutional (LF) on BLEU score, while on METEOR score, the margin increases to $3$ points. Finally, for the VANiLLa dataset, VOGUE performs considerably better compared to other baselines.

\begin{table}[t!]
\centering
\def\arraystretch{1.2}
\caption{Similarity threshold results for each dataset.}
\label{tab:sim_thres_results}
\resizebox{\textwidth}{!}{%
\begin{tabular}{lccc}
\toprule
 & \multicolumn{3}{c}{\textbf{F1-Score}} \\ \cmidrule{2-4} 
\multicolumn{1}{c}{\textbf{Module}} & \multicolumn{1}{c|}{\textbf{VQuAnDa}} & \multicolumn{1}{c|}{\textbf{ParaQA}} & \multicolumn{1}{c}{\textbf{VANiLLa}} \\ \midrule
\multicolumn{1}{l|}{Similarity Threshold} & 64.73 & 58.55 & 98.76 \\
\bottomrule
\end{tabular}%
}
\end{table}

\begin{figure*}
\centering
\includegraphics[height=0.32\textwidth]{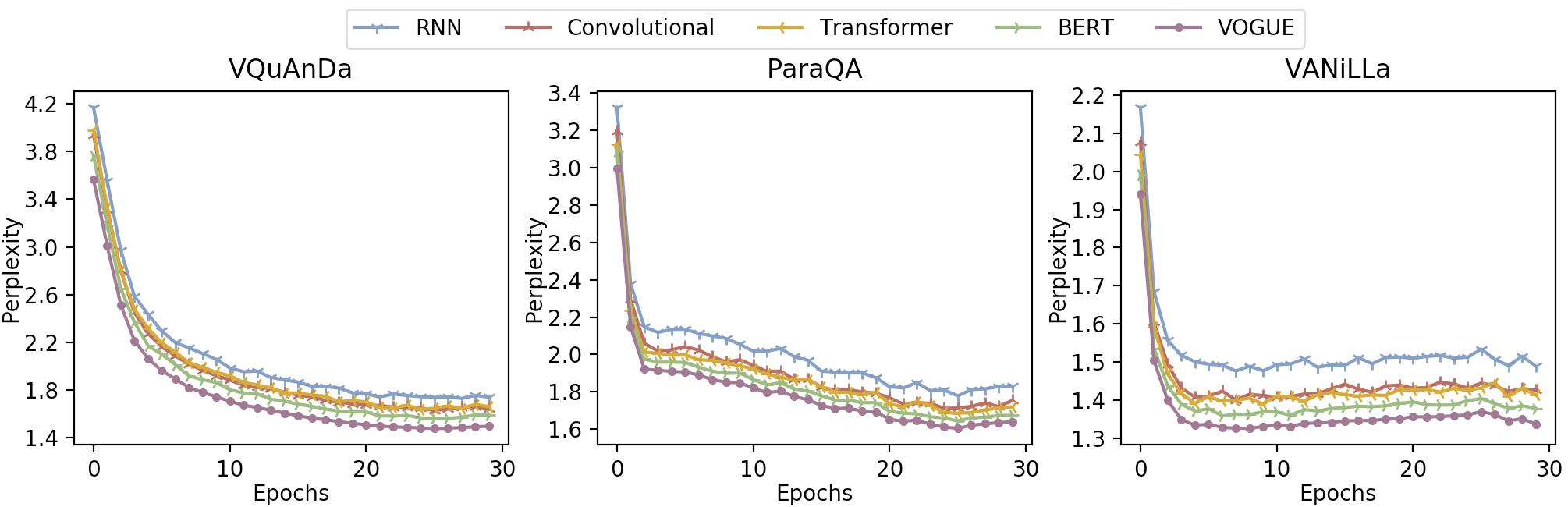}
\caption{Perplexity curves for all three answer verbalization datasets.}
\label{fig:perplexity}
\end{figure*}

\subsection{Ablation Study}

\paragraph{\textbf{Integration with Semantic Parsing based QA system}}~\\
The logical forms used for the results in Table~\ref{tab:results} are the gold ones, and therefore the performance of all baselines, including our framework, is boosted. In our first ablation study, we want to perform experiments in an end-to-end manner with a semantic parsing QA system, alongside the models, to understand our framework's superior performance. In this experiment, we train a simple, sequence-to-sequence-based semantic parser system to generate the logical forms by using the questions. As expected, the generated logical forms are not all correct, and therefore this affects the verbalization results. However, in Table~\ref{tab:results_with_qa}, we can see that VOGUE still outperforms all baselines in this setting. An important role here plays the similarity threshold module, enabling a hybrid approach even in a real-world scenario. We can only use the question as input for the baselines since we do not have the gold logical forms. Here, it is also interesting that in two of the datasets, our framework outperforms the baselines with a more significant margin than before (c.f. Table~\ref{tab:results_with_qa}, METEOR-VQuAnDa, METEOR-ParaQA).
Finally, Figure~\ref{fig:perplexity} illustrates the perplexity results, which show how well a probability distribution predicts a sample. A low perplexity indicates the probability distribution is good at predicting the sample. As we can see, our framework achieves the lowest perplexity values on all three datasets compared to other the baselines.

\paragraph{\textbf{Impact of Cross Attention and Multi-Task Learning}}~\\
Our second ablation experiment demonstrates the vitality of the cross-attention module and multi-task learning strategy. We first remove the cross-attention module from our framework. Instead, we only concatenate the question and logical form to generate the verbalization. As observed in Table~\ref{tab:ablation}, we obtain worse results compared to the original configuration of VOGUE. A simple concatenation does not interchange any information between the question and the logical form, and therefore the results are expected to be lower. The cross-attention module is intentionally built to determine relevance between inputs by jointly modeling the relationship between the question words and logical form actions. 
Next, we train all modules independently and join them on inference to understand multi-task learning efficacy. As observed, our results have a negative impact when a multi-task learning strategy is not employed.

\paragraph{\textbf{Similarity Threshold Task Analysis}}~\\
Table~\ref{tab:sim_thres_results} illustrates the performance of similarity threshold module. We observe that the module performs fairly well on VQuAnDa and ParaQA with f1 scores of $64.73$ and $58.55$, respectively. Both datasets contain complex questions. Hence, predicting the similarity between the question and the logical form is not easy. However, as long as the module's score is beyond $50$, we are confident that using the similarity threshold module can improve our frameworks' answer verbalization results.
For the VANiLLa dataset, the performance is incredibly high, with a score of $98.76$. This is because the dataset contains only simple questions. Consequently, a single template pattern is employed for this dataset, and the module here has to predict if the logical form contains the correct triple relation. The task is much easier to perform compared to complex questions. Overall, the module results are pretty accurate and encourage us to apply them in our task.

\subsection{Error Analysis}
For the error analysis, we randomly sampled 100 incorrect predictions for human evaluation. We detail the reasons for two types of errors observed in the analysis:

\paragraph{\textbf{Words Mischoose}}
A common error of VOGUE is mischoosing a word in the answer verbalization sentence. For instance, for the question ``\textit{Give me a count of everything owned by the network whose sister name is The CW?}" our framework generated the answer ``\textit{There are 156 television shows whose network's sister station is The CW.}". However, the gold reference here is ``\textit{There are 156 things whose network's sister name is The CW.}" As we can see, our framework misselected words in two parts of the sentence. The first one is the word ``\textit{things}", where it predicted ``\textit{television shows}". The second one is the word ``\textit{name}", where our model predicted ``\textit{station}". Both model predictions (``\textit{television shows}", ``\textit{station}") are correlated, since they belong to the same context. Such errors do not heavily penalize the overall performance. For the example mentioned above, the BLEU and METEOR score is positive, with values $35.74$ and $81.52$, respectively.

\paragraph{\textbf{Factual Errors}}
Another type of error of VOGUE is when it misses the semantic meaning and produces irrelevant results. It contributes to a major chunk of overall errors. There are two cases that can cause observed errors. The first one is the lack of reasoning for similar context data. When facing examples with the limited context in the dataset, the model would most definitely fail to reproduce the same context in the answer sentence. One can solve the issue by enriching the training data with other examples containing similar contexts. The second reason for having factual errors is when similarity threshold module fails to determine the inputs' relevance. As illustrated before, using the similarity threshold allows to successfully adopt a hybrid approach in a real-world scenario (\textit{QA + Answer Verbalization}) and exceed any previous baseline performance. Further, in the appendix, we describe the grammar used in this work, a case study, detailed experiment settings, and extended error analysis results.

\section{Conclusions}\label{sec:conclusion}
The considered hypothesis in the paper was to study the impact of jointly utilizing the question and logical form on the answer verbalization task. We empirically observed that the proposed ``hybrid" approach implemented in the VOGUE framework provides a flexibility to be deployed in a real world scenario where a QA system not always will produce the correct logical form.
We systematically studied the impact of our choices in the proposed architecture. For instance, the ablation study demonstrates the effectiveness of multi-task learning and cross-attention module. Albeit effective, VOGUE is the first step towards a more extensive research agenda. Based on our observations, we submit the following open research questions in this domain: 1) KGs have been recently used as a source of external knowledge in the tasks such as entity and relation linking, which are also prominent for question answering~\cite{bastos2020recon,mulang2020evaluating}. It is yet to be studied if external knowledge from KGs or other sources may positively impact the answer verbalization. 2) There are empirical evaluations that for AI systems, the explanations regarding the retrieved answers improve trustworthiness, especially in wrong prediction~\cite{kouki2017user}. Hence, how an answer verbalization can be explained remains an important open research direction. 3) In our work, we focused on English as an underlying language, and a multi-lingual approach is the next viable step. 

\bibliographystyle{acl_natbib}

\bibliography{main.bbl} 

\begin{thebibliography}{37}
\expandafter\ifx\csname natexlab\endcsname\relax\def\natexlab#1{#1}\fi

\bibitem[{Armitage et~al.(2020)Armitage, Kacupaj, Tahmasebzadeh, Swati,
  Maleshkova, Ewerth, and Lehmann}]{armitage2020mlm}
Jason Armitage, Endri Kacupaj, Golsa Tahmasebzadeh, Swati, Maria Maleshkova,
  Ralph Ewerth, and Jens Lehmann. 2020.
\newblock Mlm: A benchmark dataset for multitask learning with multiple
  languages and modalities.
\newblock In \emph{29th ACM CIKM}. ACM.

\bibitem[{Banerjee and Lavie(2005)}]{banerjee2005meteor}
Satanjeev Banerjee and Alon Lavie. 2005.
\newblock {METEOR}: An automatic metric for {MT} evaluation with improved
  correlation with human judgments.
\newblock In \emph{Proceedings of the {ACL} Workshop on Intrinsic and Extrinsic
  Evaluation Measures for Machine Translation and/or Summarization}. ACL.

\bibitem[{Bastos et~al.(2021)Bastos, Nadgeri, Singh, Mulang, Shekarpour,
  Hoffart, and Kaul}]{bastos2020recon}
Anson Bastos, Abhishek Nadgeri, Kuldeep Singh, Isaiah~Onando Mulang, Saeedeh
  Shekarpour, Johannes Hoffart, and Manohar Kaul. 2021.
\newblock Recon: Relation extraction using knowledge graph context in a graph
  neural network.
\newblock In \emph{Proceedings of The Web Conference (WWW)}, page :N/A.

\bibitem[{{Biswas} et~al.(2021){Biswas}, {Dubey}, {Rashad Al Hasan Rony}, and
  {Lehmann}}]{biswas2021vanilla}
Debanjali {Biswas}, Mohnish {Dubey}, Md~{Rashad Al Hasan Rony}, and Jens
  {Lehmann}. 2021.
\newblock \href {http://arxiv.org/abs/2105.11407} {{VANiLLa : Verbalized
  Answers in Natural Language at Large Scale}}.
\newblock \emph{arXiv e-prints}, page arXiv:2105.11407.

\bibitem[{Bordes et~al.(2017)Bordes, Boureau, and Weston}]{bordes2017god}
Antoine Bordes, Y-Lan Boureau, and Jason Weston. 2017.
\newblock Learning end-to-end goal-oriented dialog.
\newblock In \emph{5th ICLR, 2017}.

\bibitem[{{Cipolla} et~al.(2018){Cipolla}, {Gal}, and
  {Kendall}}]{cipolla2018mltloss}
R.~{Cipolla}, Y.~{Gal}, and A.~{Kendall}. 2018.
\newblock Multi-task learning using uncertainty to weigh losses for scene
  geometry and semantics.
\newblock In \emph{2018 IEEE/CVF Conference on CVPR}.

\bibitem[{Devlin et~al.(2019)Devlin, Chang, Lee, and
  Toutanova}]{devlin-etal-2019-bert}
Jacob Devlin, Ming-Wei Chang, Kenton Lee, and Kristina Toutanova. 2019.
\newblock {BERT}: Pre-training of deep bidirectional transformers for language
  understanding.
\newblock In \emph{NAACL}. ACL.

\bibitem[{Ell et~al.(2014)Ell, Harth, and Simperl}]{ell2014sparql}
Basil Ell, Andreas Harth, and Elena Simperl. 2014.
\newblock Sparql query verbalization for explaining semantic search engine
  queries.
\newblock In \emph{ESWC}.

\bibitem[{Ferr{\'e}(2017)}]{ferre2017sparklis}
S{\'e}bastien Ferr{\'e}. 2017.
\newblock Sparklis: an expressive query builder for sparql endpoints with
  guidance in natural language.
\newblock \emph{Semantic Web}.

\bibitem[{Fu et~al.(2020)Fu, Qiu, Tang, Li, Yu, and Sun}]{fu2020survey}
Bin Fu, Yunqi Qiu, Chengguang Tang, Yang Li, Haiyang Yu, and Jian Sun. 2020.
\newblock A survey on complex question answering over knowledge base: Recent
  advances and challenges.
\newblock \emph{arXiv preprint arXiv:2007.13069}.

\bibitem[{Gardent et~al.(2017)Gardent, Shimorina, Narayan, and
  Perez-Beltrachini}]{gardent-etal-2017-creating}
Claire Gardent, Anastasia Shimorina, Shashi Narayan, and Laura
  Perez-Beltrachini. 2017.
\newblock Creating training corpora for {NLG} micro-planners.
\newblock In \emph{55th ACL}. ACL.

\bibitem[{Gehring et~al.(2017)Gehring, Auli, Grangier, Yarats, and
  Dauphin}]{gehring2017conv}
Jonas Gehring, Michael Auli, David Grangier, Denis Yarats, and Yann~N. Dauphin.
  2017.
\newblock Convolutional sequence to sequence learning.
\newblock In \emph{34th ICML}, Proceedings of Machine Learning Research. PMLR.

\bibitem[{Kacupaj et~al.(2021{\natexlab{a}})Kacupaj, Banerjee, Singh, and
  Lehmann}]{kacupaj2021paraqa}
Endri Kacupaj, Barshana Banerjee, Kuldeep Singh, and Jens Lehmann.
  2021{\natexlab{a}}.
\newblock Paraqa: A question answering dataset with paraphrase responses for
  single-turn conversation.
\newblock In \emph{Eighteenth ESWC}.

\bibitem[{Kacupaj et~al.(2021{\natexlab{b}})Kacupaj, Plepi, Singh, Thakkar,
  Lehmann, and Maleshkova}]{kacupaj2021lasagne}
Endri Kacupaj, Joan Plepi, Kuldeep Singh, Harsh Thakkar, Jens Lehmann, and
  Maria Maleshkova. 2021{\natexlab{b}}.
\newblock Conversational question answering over knowledge graphs with
  transformer and graph attention networks.
\newblock In \emph{The 16th Conference of the European Chapter of the
  Association for Computational Linguistics}.

\bibitem[{Kacupaj et~al.(2020)Kacupaj, Zafar, Lehmann, and
  Maleshkova}]{kacupaj2020vquanda}
Endri Kacupaj, Hamid Zafar, Jens Lehmann, and Maria Maleshkova. 2020.
\newblock Vquanda: Verbalization question answering dataset.
\newblock In \emph{The Semantic Web}. Springer International Publishing.

\bibitem[{Kassawat et~al.(2019)Kassawat, Chaudhuri, and
  Lehmann}]{kassawat2019multi}
Firas Kassawat, Debanjan Chaudhuri, and Jens Lehmann. 2019.
\newblock Incorporating joint embeddings into goal-oriented dialogues with
  multi-task learning.
\newblock In \emph{ESWC 2019}.

\bibitem[{Kingma and Ba(2015)}]{kingma2015adam}
Diederik~P Kingma and Jimmy Ba. 2015.
\newblock Adam, a method for stochastic optimization.
\newblock In \emph{3rd ICLR}.

\bibitem[{Kouki et~al.(2017)Kouki, Schaffer, Pujara, O'Donovan, and
  Getoor}]{kouki2017user}
Pigi Kouki, James Schaffer, Jay Pujara, John O'Donovan, and Lise Getoor. 2017.
\newblock User preferences for hybrid explanations.
\newblock In \emph{Proceedings of the Eleventh ACM Conference on Recommender
  Systems}, pages 84--88.

\bibitem[{Lehmann et~al.(2015)Lehmann, Isele, Jakob, Jentzsch, Kontokostas,
  Mendes, Hellmann, Morsey, van Kleef, Auer, and Bizer}]{dbpedia}
Jens Lehmann, Robert Isele, Max Jakob, Anja Jentzsch, Dimitris Kontokostas,
  Pablo~N. Mendes, Sebastian Hellmann, Mohamed Morsey, Patrick van Kleef,
  S{\"o}ren Auer, and Christian Bizer. 2015.
\newblock Dbpedia - a large-scale, multilingual knowledge base extracted from
  wikipedia.
\newblock \emph{Semantic Web}.

\bibitem[{Lei et~al.(2018)Lei, Jin, Kan, Ren, He, and
  Yin}]{lei-etal-2018-sequicity}
Wenqiang Lei, Xisen Jin, Min-Yen Kan, Zhaochun Ren, Xiangnan He, and Dawei Yin.
  2018.
\newblock {S}equicity: Simplifying task-oriented dialogue systems with single
  sequence-to-sequence architectures.
\newblock In \emph{56th ACL}. ACL.

\bibitem[{Liu et~al.(2020)Liu, Chen, Wang, Zhang, Li, and Xu}]{ijcai2020-524}
Jie Liu, Shaowei Chen, Bingquan Wang, Jiaxin Zhang, Na~Li, and Tong Xu. 2020.
\newblock Attention as relation: Learning supervised multi-head self-attention
  for relation extraction.
\newblock In \emph{IJCAI-20}. IJCAI.

\bibitem[{Luong et~al.(2015)Luong, Pham, and Manning}]{luong2015effective}
Thang Luong, Hieu Pham, and Christopher~D. Manning. 2015.
\newblock Effective approaches to attention-based neural machine translation.
\newblock In \emph{EMNLP}. ACL.

\bibitem[{{Mohla} et~al.(2020){Mohla}, {Pande}, {Banerjee}, and
  {Chaudhuri}}]{mohla2020crossattention}
S.~{Mohla}, S.~{Pande}, B.~{Banerjee}, and S.~{Chaudhuri}. 2020.
\newblock Fusatnet: Dual attention based spectrospatial multimodal fusion
  network for hyperspectral and lidar classification.
\newblock In \emph{2020 IEEE/CVF Conference on CVPRW}.

\bibitem[{Mulang et~al.(2020)Mulang, Singh, Prabhu, Nadgeri, Hoffart, and
  Lehmann}]{mulang2020evaluating}
Isaiah~Onando Mulang, Kuldeep Singh, Chaitali Prabhu, Abhishek Nadgeri,
  Johannes Hoffart, and Jens Lehmann. 2020.
\newblock Evaluating the impact of knowledge graph context on entity
  disambiguation models.
\newblock In \emph{CIKM}.

\bibitem[{Papineni et~al.(2002)Papineni, Roukos, Ward, and
  Zhu}]{papineni2002bleu}
Kishore Papineni, Salim Roukos, Todd Ward, and Wei-Jing Zhu. 2002.
\newblock {B}leu: a method for automatic evaluation of machine translation.
\newblock In \emph{40th ACL}.

\bibitem[{Pennington et~al.(2014)Pennington, Socher, and
  Manning}]{pennington-etal-2014-glove}
Jeffrey Pennington, Richard Socher, and Christopher Manning. 2014.
\newblock {G}love: Global vectors for word representation.
\newblock In \emph{EMNLP}. ACL.

\bibitem[{Plepi et~al.(2021)Plepi, Kacupaj, Singh, Thakkar, and
  Lehmann}]{plepi2021carton}
Joan Plepi, Endri Kacupaj, Kuldeep Singh, Harsh Thakkar, and Jens Lehmann.
  2021.
\newblock Context transformer with stacked pointer networks for conversational
  question answering over knowledge graphs.
\newblock In \emph{Eighteenth ESWC}.

\bibitem[{Singh et~al.(2018)Singh, Radhakrishna, Both, Shekarpour, Lytra,
  Usbeck, Vyas, Khikmatullaev, Punjani, Lange et~al.}]{singh2018reinvent}
Kuldeep Singh, Arun~Sethupat Radhakrishna, Andreas Both, Saeedeh Shekarpour,
  Ioanna Lytra, Ricardo Usbeck, Akhilesh Vyas, Akmal Khikmatullaev, Dharmen
  Punjani, Christoph Lange, et~al. 2018.
\newblock Why reinvent the wheel: Let's build question answering systems
  together.
\newblock In \emph{Proceedings of the 2018 World Wide Web Conference}.

\bibitem[{Song et~al.(2020)Song, Wang, Su, Zhang, Xu, Ge, and
  Yu}]{song-etal-2020-structural}
Linfeng Song, Ante Wang, Jinsong Su, Yue Zhang, Kun Xu, Yubin Ge, and Dong Yu.
  2020.
\newblock Structural information preserving for graph-to-text generation.
\newblock In \emph{58th ACL}. ACL.

\bibitem[{Srivastava et~al.(2014)Srivastava, Hinton, Krizhevsky, Sutskever, and
  Salakhutdinov}]{srivastava2014dropout}
Nitish Srivastava, Geoffrey Hinton, Alex Krizhevsky, Ilya Sutskever, and Ruslan
  Salakhutdinov. 2014.
\newblock Dropout: a simple way to prevent neural networks from overfitting.
\newblock \emph{The journal of machine learning research}, 15(1):1929--1958.

\bibitem[{Unger et~al.(2012)Unger, B{\"u}hmann, Lehmann, Ngonga~Ngomo, Gerber,
  and Cimiano}]{unger2012template}
Christina Unger, Lorenz B{\"u}hmann, Jens Lehmann, Axel-Cyrille Ngonga~Ngomo,
  Daniel Gerber, and Philipp Cimiano. 2012.
\newblock Template-based question answering over rdf data.
\newblock In \emph{Proceedings of the 21st international conference on World
  Wide Web}.

\bibitem[{Vaswani et~al.(2017)Vaswani, Shazeer, Parmar, Uszkoreit, Jones,
  Gomez, Kaiser, and Polosukhin}]{vaswani2017attention}
Ashish Vaswani, Noam Shazeer, Niki Parmar, Jakob Uszkoreit, Llion Jones,
  Aidan~N. Gomez, Lukasz Kaiser, and Illia Polosukhin. 2017.
\newblock Attention is all you need.
\newblock In \emph{NIPS}.

\bibitem[{Vrande\v{c}i\'{c} and Kr\"{o}tzsch(2014)}]{wikidata}
Denny Vrande\v{c}i\'{c} and Markus Kr\"{o}tzsch. 2014.
\newblock Wikidata: A free collaborative knowledgebase.
\newblock \emph{Commun. ACM}.

\bibitem[{{Wei} et~al.(2020){Wei}, {Zhang}, {Li}, {Zhang}, and
  {Wu}}]{wei2020crossattention}
X.~{Wei}, T.~{Zhang}, Y.~{Li}, Y.~{Zhang}, and F.~{Wu}. 2020.
\newblock Multi-modality cross attention network for image and sentence
  matching.
\newblock In \emph{2020 IEEE/CVF Conference on CVPR}.

\bibitem[{Yamada et~al.(2020)Yamada, Asai, Shindo, Takeda, and
  Matsumoto}]{yamada-etal-2020-luke}
Ikuya Yamada, Akari Asai, Hiroyuki Shindo, Hideaki Takeda, and Yuji Matsumoto.
  2020.
\newblock {LUKE}: Deep contextualized entity representations with entity-aware
  self-attention.
\newblock In \emph{EMNLP}. ACL.

\bibitem[{Zhao et~al.(2020)Zhao, Walker, and
  Chaturvedi}]{zhao-etal-2020-bridging}
Chao Zhao, Marilyn Walker, and Snigdha Chaturvedi. 2020.
\newblock Bridging the structural gap between encoding and decoding for
  data-to-text generation.
\newblock In \emph{58th ACL}. ACL.

\bibitem[{Zheng et~al.(2017)Zheng, Cheng, Zou, Yu, and Zhao}]{zheng2017natural}
Weiguo Zheng, Hong Cheng, Lei Zou, Jeffrey~Xu Yu, and Kangfei Zhao. 2017.
\newblock Natural language question/answering: Let users talk with the
  knowledge graph.
\newblock In \emph{2017 ACM CIKM}.

\end{thebibliography}
\appendix

\begin{table*}[ht!]
\centering
\caption{Dataset examples annotated with gold logical forms.}
\label{tab:dataset_examples}
\def\arraystretch{1.2}
\begin{tabular}{lll}
\toprule
\multicolumn{1}{c}{\textbf{Dataset}} & \textbf{Question} & \textbf{Logical Form} \\ \midrule
\multirow{5}{*}{VQuAnDa} & \begin{tabular}[c]{@{}l@{}}In which team did Dave Bing and\\ Ron Reed started their basketball career?\end{tabular} & \begin{tabular}[c]{@{}l@{}}union(find(Dave\_Bing, draftteam),\\ find(Ron\_Reed, draftteam))\end{tabular} \\ \cline{2-3} 
 & \begin{tabular}[c]{@{}l@{}}Does the white river flow into\\ the connecticut river?\end{tabular} & \begin{tabular}[c]{@{}l@{}}is\_in(find(Connecticut\_River, rightTributary),\\ White\_River\_(Vermont))\end{tabular} \\ \cline{2-3} 
 & \begin{tabular}[c]{@{}l@{}}Which sports are played in schools affiliated\\ with the Harvest Christian Center?\end{tabular} & \begin{tabular}[c]{@{}l@{}}find(filter\_type(find\_reverse(Harvest\_Christian\_Center,\\ religiousAffiliation), School), sport)\end{tabular} \\ \midrule
\multirow{6}{*}{ParaQA} & \begin{tabular}[c]{@{}l@{}}How many people have been part of\\ Chicago Bulls team?\end{tabular} & \begin{tabular}[c]{@{}l@{}}count(find\_reverse(Chicago\_Bulls, team))\end{tabular} \\ \cline{2-3} 
 & \begin{tabular}[c]{@{}l@{}}Name the office holder whose alma mater\\ is Harvard-Westlake School and resting place\\ is Alta Mesa Memorial Park?\end{tabular} & \begin{tabular}[c]{@{}l@{}}intersection(filter\_type(find\_reverse(\\Harvard-Westlake\_School, almamater), officeholder),\\ find\_reverse(Alta\_Mesa\_Memorial\_Park, restingplace))\end{tabular} \\ \cline{2-3} 
 & \begin{tabular}[c]{@{}l@{}}Name all the broadcast area of the TV stations\\ which has Rodrigues as one of the broadcast area?\end{tabular} & \begin{tabular}[c]{@{}l@{}}find(filter\_type(find\_reverse(Rodrigues, broadcastArea),\\ TelevisionStation), broadcastArea)\end{tabular} \\ \midrule
\multirow{5}{*}{VANiLLa} & \begin{tabular}[c]{@{}l@{}}What is the nonprofit organization\\ where Bruce Bochy was educated?\end{tabular} & \begin{tabular}[c]{@{}l@{}}filter\_type(find(Q586449, P69), Q163740)\end{tabular} \\ \cline{2-3} 
 & \begin{tabular}[c]{@{}l@{}}Which language can Paolo Brera\\ understand?\end{tabular} & \begin{tabular}[c]{@{}l@{}}find(Q2423068, P1412)\end{tabular} \\ \cline{2-3} 
 & \begin{tabular}[c]{@{}l@{}}Which administrative territory is Shaun\\ Cunnington an inhabitant of ?\end{tabular} & \begin{tabular}[c]{@{}l@{}}filter\_type(find(Q7490823, P27))\end{tabular} \\
 \bottomrule
\end{tabular}
\end{table*}

\section{Appendix}
Due to page limit, we could not put several empirical results in the main paper. This section describes the remaining empirical studies. 

\section{Grammar}
For the logical forms, we employ a grammar that can be used to capture the entire context of the question with the minimum number of actions. We prefer not to reinvent the wheel, and therefore we adopted the grammar from existing state-of-the-art question answering systems~\cite{kacupaj2021lasagne,plepi2021carton}. However, we do not employ all the actions from these works; Table~\ref {tab:grammar} illustrates the complete grammar with all the defined actions that we used for all three answer verbalization datasets. As we can see, for a couple of actions, we also have their reverse occurrence (e.g. \textit{find}, \textit{find\_reverse}).
This is done to match the knowledge graph triple direction (\textit{subject-predicate-object}). In some questions, we might have the subject or the object entity. Having both normal and reverse actions helps us identify the correct answer directly based on the model's predicted action. In Table~\ref{tab:dataset_examples}, we illustrate how the actions can be used to annotate questions from all three datasets.
For instance, for the VQuAnDa question ``\textit{Which sports are played in schools affiliated with the Harvest Christian Center?}" the gold logical form does include three different actions from our grammar (\textit{find}, \textit{filter\_type} and \textit{find\_reverse}). Where the ``\textit{find\_reverse}" is used to identify all the subject entities of the triple (\textit{?subject}, \textit{religiousAffiliation}, \textit{Harvest\_Christian\_Center}). Another interesting example is the ParaQA question ``\textit{Name the office holder whose alma mater is Harvard-Westlake School and resting place is Alta Mesa Memorial Park?}" where the gold logical form contains the action ``\textit{intersection}" which allows us to identify the intersection between two sets of entities. Similarly, we also use the ``\textit{union}" action. Next, for the quantitative questions, we employ the action ``\textit{count}" which returns the length of the entity set. Finally, for verification questions, we have the ``\textit{is\_in}" action, which checks whether an entity exists in a set. 

\begin{table}[t!]
\centering
\def\arraystretch{1.1}
\caption{Predefined grammar with respective actions to generate logical forms.}
\label{tab:grammar}
\resizebox{\textwidth}{!}{%
\begin{tabular}{ll}
\toprule
\textbf{Action} & \textbf{Description} \\
\midrule
set $\rightarrow$ find(\textit{e}, \textit{p}) & \begin{tabular}[c]{@{}l@{}}set of objects part of the triples\\ with subject \textit{e} and predicate \textit{p}\end{tabular} \\
set $\rightarrow$ find\_reverse(\textit{e}, \textit{p}) & \begin{tabular}[c]{@{}l@{}}set of subjects part of the triples\\ with object \textit{e} and predicate \textit{p}\end{tabular} \\
set $\rightarrow$ filter\_type(set, tp) & \begin{tabular}[c]{@{}l@{}}filter the given set of entities\\ based on the given type\end{tabular} \\
boolean $\rightarrow$ is\_in(entity, set) & check if the entity is in the set  \\
number $\rightarrow$ count(set) & number of elements in the set \\
set $\rightarrow$ union($set_1$, $set_2$) & union of $set_1$ and $set_2$\\
set $\rightarrow$ intersection($set_1$, $set_2$) & intersection of $set_1$ and $set_2$ \\
\bottomrule
\end{tabular}%
}
\end{table}

\section{Case Study}
We further manually inspect our framework VOGUE outputs for conducting a case study to understand the performance better. As shown in Table~\ref{tab:case_study}, we find that in the first example, VOGUE can produce the exact verbalization with the reference. Here the logical form contains three actions (\textit{count} and two \textit{find\_reverse}), and is not a simple question. Such results indicate the superb performance of our framework. Next, we can see the question ``\textit{Who is the scientist whose academic advisor was Karl Ewald Hasse?}" here, VOGUE manages to generate almost the exact verbalization. In particular, it only mischoses a single word, which is still relevant to the context, and the generated result could be easily considered correct. Our framework here generated the answer ``\textit{The scientist whose doctoral advisor is Karl Ewald Hasse is Robert Koch.}" and the word it missed was ``\textit{academic}" where it replaced it with ``\textit{doctoral}". The following example illustrates the robustness of our framework. Here the question is ``\textit{What are the movies with Daniel Waters as screenwriter?}" and our model produces a flawless verbalization and it only complicates the words ``\textit{screenwriter}" and ``\textit{director}". It also replaces the word ``\textit{films}" with ``\textit{movies}" which can be considered synonyms and make no significant difference in verbalization. The generated response here is grammatically sound and adequately represent all the information in the question and logical form. Finally, in the last example, we can see that VOGUE has generalized in the answer verbalization compared to the reference. More precisely, the question here refers to cars designed by a company, and the reference also mentions it. However, our model produces a more general but at the same time fluent verbalization that refers to ``\textit{things}" instead of ``\textit{cars}", which again can be considered a valid response for the given input.

\section{Hyperparameters and Module Configurations}
Table~\ref{tab:hyper} summarizes the hyperparameters used across the VOGUE framework. Starting with training parameters, we employ a batch size of $256$, a learning rate of $0.001$ and we train for $100$ epochs. For the optimization, we use the Noam optimizer proposed by \cite{vaswani2017attention}, where authors use an Adam optimizer \cite{kingma2015adam} with several warmup steps for the learning rate. In our case, the number of warmup steps is $4000$. During optimization, we clip the gradients with a max norm of $5$. We apply a dropout with a probability $0.1$ athwart our framework and use an embedding dimension of $d=300$. All our modules operate under the same embedding dimension. We apply the GloVe word embedding model to our input tokens with a word embedding dimension of $300$. For the transformer encoder and decoder, we use the configurations from \cite{vaswani2017attention}. Our model dimension is $d=300$, with a total number of $H=6$ heads and $L=2$ layers. The inner feed-forward linear layers have dimension $d_{ff}  = 600$, (2 * 300). Following the base transformer parameters, we apply residual dropout \cite{srivastava2014dropout} to the summation of the embeddings and the positional encodings in both encoder and decoder stacks with a rate of $0.1$. 
The similarity threshold module receives an input of dimension $600$ where here a linear layer is responsible for reducing it to $300$, which is the framework dimension. Next, a LeakyReLU, dropout, and a linear layer are used for the final prediction. 
Finally, for the cross attention module, we apply hyperparameters similar to the transformer model. Our dimension remains of size $d=300$ and again the number of heads is $H=6$. However, we do not apply multiple layers here. Likewise, dropout is applied with probability $0.1$. The number of training parameters for VQuAnDa, ParaQA, and VANiLLa datasets are 12.9M, 14.9M, and 46.8M respectively. 

\begin{table}[t!]
\centering
\caption{Hyperparameters for VOGUE framework.}
\label{tab:hyper}
\begin{tabular}{cc}
\toprule
\textbf{Hyperparameters} & \textbf{Value} \\ \midrule
epochs & 100 \\
batch size & 256 \\
learning rate & 0.001 \\
dropout ratio & 0.1 \\
optimizer & Adam \\
warmup steps & 4000 \\
clip max norm & 5 \\
$\beta_1$ & 0.9 \\
$\beta_2$ & 0.999 \\
$\varepsilon$ & 1e-09 \\
model dim & 300 \\
word embedding model & GloVe \\
word embedding dim & 300 \\
transformer layers & 2 \\
transformer heads & 6 \\
\begin{tabular}[c]{@{}l@{}}similarity threshold\\ non-linear activation\end{tabular} & LeakyRelu \\
\bottomrule
\end{tabular}
\end{table}

\section{Detailed Experiments}
We provide detailed experiment results for metrics such as perplexity, BLEU-1, BLEU-2, BLEU-3, BLEU-4 and METEOR in the following Tables~\ref{tab:vquanda_results},~\ref{tab:paraqa_results},~\ref{tab:vanilla_results}.

\begin{table*}[ht!]
\centering
\caption{Sample output of our framework.}
\label{tab:case_study}
\def\arraystretch{1.2}
\begin{tabular}{ll}
\toprule
Question & \begin{tabular}[c]{@{}l@{}}How many other home stadium are there of the soccer club \\ whose home stadium is Luzhniki Stadium?\end{tabular} \\ \cline{2-2}
Logical Form & \begin{tabular}[c]{@{}l@{}}count(find\_reverse(find\_reverse(Luzhniki\_Stadium, homeStadium), \\ homeStadium))\end{tabular} \\ \cline{2-2}
Reference & \begin{tabular}[c]{@{}l@{}}There are 9 home stadiums of the soccer club \\ whose home stadium is Luzhniki Stadium.\end{tabular} \\ \cline{2-2}
VOGUE & \begin{tabular}[c]{@{}l@{}}There are 9 home stadiums of the soccer club \\ whose home stadium is Luzhniki Stadium.\end{tabular} \\ \midrule
Question & Who is the scientist whose academic advisor was Karl Ewald Hasse? \\ \cline{2-2}
Logical Form & \begin{tabular}[c]{@{}l@{}}filter\_type(find\_reverse(Karl\_Ewald\_Hasse, academicAdvisor), \\ Scientist)\end{tabular} \\ \cline{2-2}
Reference & \begin{tabular}[c]{@{}l@{}}The scientist whose \textbf{academic advisor} is Karl Ewald Hasse \\ is Robert Koch.\end{tabular} \\ \cline{2-2}
VOGUE & \begin{tabular}[c]{@{}l@{}}The scientist whose \underline{doctoral advisor} is Karl Ewald Hasse \\ is Robert Koch.\end{tabular} \\ \midrule
Question & What are the movies with Daniel Waters as screenwriter? \\ \cline{2-2}
Logical Form & filter\_type(find\_reverse(Daniel\_Waters, screenplay), Film) \\ \cline{2-2}
Reference & \begin{tabular}[c]{@{}l@{}}The \textbf{films} with the \textbf{screenplay written} by Daniel Waters are Batman Returns, \\ Demolition Man (film), Hudson Hawk, The Adventures of Ford Fairlane.\end{tabular} \\ \cline{2-2}
VOGUE & \begin{tabular}[c]{@{}l@{}}The \underline{movies} whose \underline{director} is Daniel Waters are Batman Returns, \\ Demolition Man (film), Hudson Hawk, The Adventures of Ford Fairlane.\end{tabular} \\ \midrule
Question & \begin{tabular}[c]{@{}l@{}}Which person designed the cars which has been designed by \\ ASC Creative Services?\end{tabular} \\ \cline{2-2}
Logical Form & \begin{tabular}[c]{@{}l@{}}find(filter\_type(find\_reverse(ASC\_Creative\_Services, designCompany), \\ Automobile), designer)\end{tabular} \\ \cline{2-2}
Reference & \begin{tabular}[c]{@{}l@{}}The \textbf{designers} of the \textbf{cars} whose designer company is ASC Creative \\ Services are Warren, Michigan, Michigan, ASC Creative Services.\end{tabular} \\ \cline{2-2}
VOGUE & \begin{tabular}[c]{@{}l@{}}The \underline{things} which have been \underline{designed} by ASC Creative Services are \\ Warren, Michigan, Michigan, ASC Creative Services.\end{tabular} \\
\bottomrule
\end{tabular}
\end{table*}

\clearpage

\begin{table*}[ht!]
\centering
\caption{Detailed results on VQuAnDa dataset.}
\label{tab:vquanda_results}
\begin{tabular}{lcccccc}
\toprule
 & \multicolumn{6}{c}{\textbf{VQuAnDa}} \\ \cmidrule{2-7}
\textbf{Models} & \multicolumn{1}{l|}{\textbf{Perplexity}} & \multicolumn{1}{l|}{\textbf{BLEU-1}} & \multicolumn{1}{l|}{\textbf{BLEU-2}} & \multicolumn{1}{l|}{\textbf{BLEU-3}} & \multicolumn{1}{l|}{\textbf{BLEU-4}} & \multicolumn{1}{l}{\textbf{METEOR}} \\ \midrule
\multicolumn{1}{l|}{RNN (\textbf{Q})} & 1.8132 & 50.71 & 35.11 & 21.65 & 15.43 & 53.15 \\
\multicolumn{1}{l|}{RNN (\textbf{LF})} & 1.7297 & 54.38 & 40.17 & 23.12 & 20.19 & 57.06 \\
\multicolumn{1}{l|}{Convolutional (\textbf{Q})} & 1.6696 & 58.49 & 42.86 & 27.87 & 21.32 & 57.54 \\
\multicolumn{1}{l|}{Convolutional (\textbf{LF})} & 1.6170 & 62.58 & 44.79 & 30.43 & 26.02 & 64.30 \\
\multicolumn{1}{l|}{Transformer (\textbf{Q})} & 1.6953 & 57.27 & 40.22 & 26.92 & 18.37 & 56.83 \\
\multicolumn{1}{l|}{Transformer (\textbf{LF})} & 1.6434 & 60.34 & 44.38 & 31.45 & 23.18 & 60.17 \\
\multicolumn{1}{l|}{BERT (\textbf{Q})} & 1.6474 & 59.78 & 43.64 & 28.25 & 20.78 & 59.28 \\
\multicolumn{1}{l|}{BERT (\textbf{LF})} & 1.5575 & 62.20 & 45.27 & 32.83 & 24.96 & 65.92 \\ \midrule
\multicolumn{1}{l|}{VOGUE (ours) (\textbf{H})} & \textbf{1.4791} & \textbf{65.97} & \textbf{49.20} & \textbf{36.71} & \textbf{28.76} & \textbf{67.21} \\
\bottomrule
\end{tabular}
\end{table*}

\begin{table*}[ht!]
\centering
\caption{Detailed results on ParaQA dataset.}
\label{tab:paraqa_results}
\begin{tabular}{lcccccc}
\toprule
 & \multicolumn{6}{c}{\textbf{ParaQA}} \\ \cmidrule{2-7}
\textbf{Models} & \multicolumn{1}{l|}{\textbf{Perplexity}} & \multicolumn{1}{l|}{\textbf{BLEU-1}} & \multicolumn{1}{l|}{\textbf{BLEU-2}} & \multicolumn{1}{l|}{\textbf{BLEU-3}} & \multicolumn{1}{l|}{\textbf{BLEU-4}} & \multicolumn{1}{l}{\textbf{METEOR}} \\ \midrule
\multicolumn{1}{l|}{RNN (\textbf{Q})} & 1.8816 & 58.56 & 40.64 & 28.74 & 22.45 & 58.41 \\
\multicolumn{1}{l|}{RNN (\textbf{LF})} & 1.7768 & 61.87 & 44.48 & 31.86 & 26.36 & 61.87 \\
\multicolumn{1}{l|}{Convolutional (\textbf{Q})} & 1.7879 & 60.15 & 43.29 & 30.15 & 25.94 & 60.82 \\
\multicolumn{1}{l|}{Convolutional (\textbf{LF})} & 1.7056 & 65.93 & 49.95 & 37.98 & 31.89 & 65.85 \\
\multicolumn{1}{l|}{Transformer (\textbf{Q})} & 1.7701 & 59.68 & 42.22 & 30.33 & 23.61 & 59.63 \\
\multicolumn{1}{l|}{Transformer (\textbf{LF})} & 1.6731 & 64.04 & 47.73 & 36.62 & 28.01 & 63.75 \\
\multicolumn{1}{l|}{BERT (\textbf{Q})} & 1.7217 & 61.21 & 45.69 & 32.29 & 24.12 & 62.59 \\
\multicolumn{1}{l|}{BERT (\textbf{LF})} & 1.6459 & 66.47 & 50.91 & 37.61 & 29.31 & 65.92 \\ \midrule
\multicolumn{1}{l|}{VOGUE (ours) (\textbf{H})} & \textbf{1.6043} & \textbf{67.60} & \textbf{51.94} & \textbf{39.94} & \textbf{32.05} & \textbf{68.85} \\
\bottomrule
\end{tabular}
\end{table*}

\begin{table*}[ht!]
\centering
\caption{Detailed results on VANiLLa dataset.}
\label{tab:vanilla_results}
\begin{tabular}{lcccccc}
\toprule
 & \multicolumn{6}{c}{\textbf{VANiLLa}} \\ \cmidrule{2-7}
\textbf{Models} & \multicolumn{1}{l|}{\textbf{Perplexity}} & \multicolumn{1}{l|}{\textbf{BLEU-1}} & \multicolumn{1}{l|}{\textbf{BLEU-2}} & \multicolumn{1}{l|}{\textbf{BLEU-3}} & \multicolumn{1}{l|}{\textbf{BLEU-4}} & \multicolumn{1}{l}{\textbf{METEOR}} \\ \midrule
\multicolumn{1}{l|}{RNN (Q)} & 1.5629 & 41.29 & 30.85 & 22.76 & 16.66 & 58.67 \\
\multicolumn{1}{l|}{RNN (LF)} & 1.4708 & 42.33 & 29.71 & 23.16 & 16.45 & 55.34 \\
\multicolumn{1}{l|}{Convolutional (Q)} & 1.4715 & 40.45 & 28.39 & 23.38 & 15.42 & 61.14 \\
\multicolumn{1}{l|}{Convolutional (LF)} & 1.4043 & 42.10 & 29.53 & 22.34 & 16.89 & 58.72 \\
\multicolumn{1}{l|}{Transformer (Q)} & 1.4683 & 55.09 & 40.77 & 35.62 & 30.80 & 62.16 \\
\multicolumn{1}{l|}{Transformer (LF)} & 1.3817 & 53.37 & 39.28 & 34.98 & 28.12 & 59.01 \\
\multicolumn{1}{l|}{BERT (Q)} & 1.4193 & 57.23 & 43.83 & 37.47 & 31.32 & 62.96 \\
\multicolumn{1}{l|}{BERT (LF)} & 1.3557 & 56.03 & 42.59 & 36.22 & 30.11 & 59.27 \\ \midrule
\multicolumn{1}{l|}{VOGUE (ours) (H)} & \textbf{1.3261} & \textbf{62.72} & \textbf{48.77} & \textbf{40.80} & \textbf{35.14} & \textbf{65.04} \\
\bottomrule
\end{tabular}
\end{table*}
\end{document}